\documentclass[conference]{IEEEtran}
\IEEEoverridecommandlockouts
\usepackage{cite}
\usepackage{scrextend}
\usepackage{amsmath,amssymb,amsfonts}
\usepackage{algorithmic}
\usepackage{graphicx}
\usepackage{textcomp}
\usepackage{xcolor}
\usepackage{tikz}
\usetikzlibrary{positioning}
\usetikzlibrary{shapes, arrows, calc, fit}

\def\BibTeX{{\rm B\kern-.05em{\sc i\kern-.025em b}\kern-.08em
    T\kern-.1667em\lower.7ex\hbox{E}\kern-.125emX}}
\begin{document}

\title{A Systems Theoretic Approach to Online Machine Learning}

\author{Anli du Preez,
        Peter Beling,
        Tyler Cody$\IEEEauthorrefmark{1}$ \\
        \small Grado Department of Industrial and Systems Engineering, Virginia Tech, Blacksburg, VA, USA \\
        \small Responsible General Intelligence Lab, Virginia Tech, Arlington, VA, USA \\
        \small $\IEEEauthorrefmark{1}$Corresponding Author: tcody@vt.edu \\
}

\maketitle

\begin{abstract}
The machine learning formulation of online learning is incomplete from a systems theoretic perspective. Typically, machine learning research emphasizes domains and tasks, and a problem solving worldview. It focuses on algorithm parameters, features, and samples, and neglects the perspective offered by considering system structure and system behavior or dynamics. Online learning is an active field of research and has been widely explored in terms of statistical theory and computational algorithms, however, in general, the literature still lacks formal system theoretical frameworks for modeling online learning systems and resolving systems-related concept drift issues. Furthermore, while the machine learning formulation serves to classify methods and literature, the systems theoretic formulation presented herein serves to provide a framework for the top-down design of online learning systems, including a novel definition of online learning and the identification of key design parameters. The framework is formulated in terms of input-output systems and is further divided into system structure and system behavior. Concept drift is a critical challenge faced in online learning, and this work formally approaches it as part of the system behavior characteristics. Healthcare provider fraud detection using machine learning is used as a case study throughout the paper to ground the discussion in a real-world online learning challenge. 
\end{abstract}

\begin{IEEEkeywords}
machine learning, online learning, systems theory, learning theory
\end{IEEEkeywords}

\section{Introduction}
Unlike typical machine learning (ML) methods that assume training examples are available before the learning task, online learning (OL) addresses the challenge of most real-world problems where training data arrives sequentially over time \cite{b1}. OL is a method of ML where the goal of the learner is to sequentially update itself when presented with a sequential stream of data in order to remain the best predictor for future data at every time step \cite{b2}. Thus, the learner seeks to accumulate knowledge and update it according to the changes in the data distribution over time to maintain reliable prediction performance. OL is relevant to all types of learning and is most widely studied in the context of supervised learning, where full feedback information (i.e., labeled data) is assumed to always be available to learn from. 

In this paper, a systems theoretic framework is developed to aid in the systems understanding of OL algorithms and the associated knowledge updating process in terms of systems terminology. The presented systems theoretic approach allows for the top-down design of OL systems, because it considers OL systems in terms of their abstract, general-systems nature, as opposed to the specific details of OL solution methods. Furthermore, a novel definition of OL as well as the identification of key algorithm design parameters in OL systems are provided. 

Throughout the paper, healthcare provider fraud detection via ML will be used as a running example to ground the theoretical discussion. In healthcare fraud, the need for OL arises in numerous scenarios. Fraudsters adjust their behavior over time to remain successful and undetected, hence the data distributions change over time and the applied learner should update accordingly to remain relevant and trustworthy. Healthcare providers can change a variety of factors, such as their billing prices and procedures, prescription behavior, and treatments administered, to disguise their fraud. It is expected that the fraud detection learning system will remember previous fraudulent behavior to identify the fraud when it occurs again, as well as to be on guard to detect new forms of fraud, learn from it, and incorporate it into its detection mechanism for future applications.

Despite a broad literature of solution methods, existing surveys on OL and concept drift (CD) do not provide concepts for OL systems themselves that rise to the level of formalism provided herein \cite{b2, b8, CDsurvey1, CDsurvey2}. And while there are more formal results for OL in the learning theory literature \cite{OLtheory1, OLtheory2}, they are largely statistical, not systems theoretic. Regarding systems theoretic approaches, abstract learning systems theory (ALST) has been proposed as a systems theoretic approach to studying learning systems, following the formal-minimalist approach to general systems theory of Mesarovic \cite{wv, b21, b22}. While it has been applied to statistical learning theory, transfer learning, meta-learning, and multi-task learning \cite{cody1, hpaper, cody2}, it has not been extended to OL. Thus, the definitions and concepts provided by this paper extend the OL and CD literature with formalism, the learning theory literature with systems theory, and the learning systems theory literature with the specifics of OL.

This paper is organized as follows. First, a background on both OL and healthcare fraud detection is provided. Then, the ML formulation of OL is reviewed. Next, the novel definition of OL systems is presented, followed by an in-depth discussion of the system structure and system behavior of OL systems. Lastly, the paper is concludes with remarks on the limitations of the presented framework as well as directions for future research.

\section{Background}

\subsection{Online Learning Background}

ML is becoming increasingly popular in modern-day applications of data analysis and artificial intelligence (AI). Traditionally, ML concerned three types of learning: unsupervised, semi-supervised, and supervised learning, however, ML concepts have since been developed and expanded to include a variety of learning types, including reinforcement, meta-, multi-task, transfer, online, and active learning.

Traditional ML methods are often referred to as offline ML methods. In offline ML settings, all the data that will be available to train (``learn'') ML models is available all at once, prior to model deployment. The model is usually deployed for inference or predictive tasks without (or rarely) performing routine updates to the learned model over its life cycle. Thus, there is an inherent assumption that the training and deployment environments will be nearly identical. But since real-world conditions are continually evolving, the distributions of the data they generate are ever changing. As a result, offline ML methods are poorly equipped to adapt and require expensive re-training when new training data (from a new distribution) is presented to the model since, in principle, an entirely new model has to be learned. Hence, an additional drawback of traditional offline ML methods is their low efficiency in terms of both time and space costs---retraining models from scratch takes time and requires space to store collected data.

A potential solution to these challenges is OL, a ML method where the model learns incrementally from a sequence of data instances one by one (or in mini-batches) at each time step. Unlike traditional ML, OL assumes that the initial and continuing conditions of the operating environment may not be or remain the same over time. The goal of OL is to maximize the model's accuracy or correctness for the sequence of predictions made by the learned model, given the correct answers to previous prediction tasks and possibly additional information. The model structure is designed such that the learned model can be updated in near real-time as new training data arrives, addressing the major time and space costs of traditional offline ML methods. Thus, OL methods are more efficient and scalable for large-scale ML tasks in real-world data analysis applications where data arrives in high volume, at high velocity, and with varying data distributions  \cite{b2}.

\subsection{Healthcare Fraud and Online Learning}

Many real-world applications for data analysis concern non-stationary environments, where the underlying data distribution changes over time, sometimes termed CD, such as in climate or financial data analysis, information retrieval, web mining, network intrusion, spam and fraud detection, and elsewhere \cite{b3}. Fraud, for example, takes many shapes and forms, including, credit card, healthcare, and insurance fraud.  The National Health Care Anti-Fraud Association (NCHAA) in the United States estimates that health insurance fraud cost the nation approximately 123-410 billion USD for the year of 2021 \cite{b4}. Provider fraud is the most prominent healthcare fraud that occurs. Providers can change a variety of factors, such as their billing prices and procedures, prescription behavior, and treatments administered, to disguise their fraud. Not only is there continuous change in the data distribution over time, but billions of healthcare transactions occur annually, hence healthcare data are big data with high velocity and volume.


Healthcare fraud detection systems require scalable algorithms that are low-cost and efficient in both space and time requirements. In conclusion, healthcare fraud is an extremely expensive challenge with attributes suitable for OL applications.


\subsection{Abstract Learning Systems Theory }

Systems theory is a branch of philosophy and mathematics that states that the component parts of a system can best be understood in the context of their relationships with each other and with other systems rather than in isolation. Formal systems theory utilizes mathematics to make formal, meticulous statements about systems. The systems theoretic framework developed in this paper has foundations in the abstract systems theory (AST) proposed by Mesarovic and Takahara \cite{b21, b22}.

In (AST), a system $S$ is defined as a proper relation on component sets, called system objects. When the set of system's objects can be partitioned into two classes, namely an input $X$ and output $Y$ set, the system is called an input–output or terminal system. This system is described by $S \subset X \times Y$. The input set consists of the objects (sets) representing the influence from the environment, and the output set consists of objects (sets) representing the influence from the system on the environment. Input-output systems are called functional systems when $S: X \to Y$. The system structure for an input-output system is defined by the sample spaces $X$ and $Y$, hence the observed data. The system behavior is defined by measures on the function $S(X)=Y$.

Recently, abstract learning systems theory (ALST) has been proposed as an extension of AST to learning systems that serves as a meta-theory for ML and learning theory \cite{cody1, cody2, hpaper, cody2020motivating, cody2023cascading}. In this paper, ALST is used to formulate and study OL systems. In ALST, a learning system is described in terms of the definition of an input-output system. A learning system $S_L$ uses the data $D \subset X \times Y$ to determine a predictive function $f : X \rightarrow Y$ using inference. Implicitly, the learning system consists of an algorithm $A:D \to \Theta$ that interacts with the  data $D$ to learn model parameters $\Theta$ that govern the (parameterized) predictive function $f: \Theta \times X \to Y$, which maps the parameters and input to the output. At the input-output level, the learning system is a relation $S\subset\times\{D, X, Y\}$. Figure \ref{fig:inout} illustrates both an input-output system and learning as an input-output system.

\tikzstyle{block} = [draw, rectangle, minimum height=2em, minimum width=2em]
\tikzstyle{sum} = [draw, circle, node distance=1cm]
\tikzstyle{input} = [coordinate]
\tikzstyle{output} = [coordinate]
\tikzstyle{block1} = [draw, rectangle, minimum height=4em, minimum width=4em]

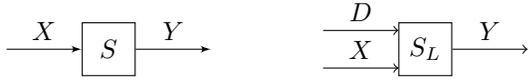
\begin{figure}
\centering
\begin{tikzpicture}[auto, node distance=2cm,>=latex']
\node [input, name=input] {};
\node [block, right=1cm of input] (system) {$S$};
\node [output, right=1cm of system] (output) {};
\draw [draw,->] (input) -- node{$X$} (system);
\draw [->] (system) -- node{$Y$}(output);
\end{tikzpicture}
\hspace{1cm}
\begin{tikzpicture}
\node (input){};
\node [block, right = 1cm of input, name=g]{$S_L$};
\node [output, right =1cm of g, name=output]{};

    \begin{scope}[transform canvas={yshift=+2.5mm}]
        \draw [->] (input) -- (g) node [midway, above] {$D$};
    \end{scope}
    \begin{scope}[transform canvas={yshift=-2.5mm}]
        \draw [->] (input) -- (g) node [midway, above] {$X$};
    \end{scope}
    
    \begin{scope}
        \draw [->] (g) -- (output) node [midway, above] {$Y$};
    \end{scope}

\end{tikzpicture}
\caption{An input-output system $S$ (left) and learning as an input-output system $S_L$ (right).}
\label{fig:inout}
\end{figure}

Note that $D \subset X \times Y$ by convention, i.e., in accordance with supervised learning, where it is assumed that the cardinality of $X$ is the same as for $Y$. In terms of semi-supervised learning, $D \subset X \times Y$ still holds, however, the cardinality of $X$ is larger than that of $Y$. Alternatively, the data $D$ could be a subset of the input only, i.e., $D \subset X$, as in unsupervised learning settings. 

\noindent \\ \textbf{Example II.1: Learning System for Healthcare}\\
\noindent Throughout this paper a healthcare fraud detection learning system will be considered. The input data $X$ consists of healthcare provider claims that are submitted to a healthcare insurer for reimbursement over time. The continuous online learner $S$ uses the input data to infer the predictive function $f$, which is used to predict the output data or labels $Y$ over time. Generally, fraud detection is approached as a binary classification task of predicting labels $y\in Y$ of ``normal" or ``fraudulent", however, it can be extended to a multi-class classification task to aid in identifying specific types of fraudulent behavior.

\section{Online Learning}
OL is an intuitive concept widely studied by computer scientists specializing in ML \cite{b1,b2,b5, b6, b7, b8, b9, b10 }. The generality of the OL framework makes it a super-structure for many learning problems. It is closely related to transfer learning \cite{cody1}, reinforcement learning \cite{b1}, domain adaptation \cite{b11, b12}, and CD \cite{b7, b13, b14, b15}, which focus on sharing data and accumulating knowledge to improve learning, and loosely related to meta-learning and learning to learn \cite{b16,b17,b18,b19,b20}, which focus on improving the learning process itself.

\subsection{Approaches in Online Learning}
According to the types of feedback information and the types of supervision in the learning tasks, OL techniques can be classified into the following three major categories: OL with full feedback, OL with limited feedback, and online unsupervised learning \cite{b2}. In OL with full feedback, information is always revealed to a learner at the end of each OL round. In OL with limited feedback, the learner receives partial feedback information from the environment during OL. For such tasks, the learner often has to make online updates or decisions by attempting to achieve a trade-off between the exploitation of the provided feedback and the exploration of unknown information in the environment. Online unsupervised learning can be considered as an extension of traditional unsupervised learning (the learner receives data without any additional feedback, for example, true class labels) for dealing with continuous data streams. 

\subsection{Challenges in Online Learning}
Even though extensive studies have been conducted in the literature, many unresolved open challenges in OL still remain. Three high-level issues exist, namely: addressing large-scale, real-time big data analysis, the continued exploration of innovative OL algorithms, and the critical challenge of CD in OL \cite{b2}.

CD describes the setting where the distribution of data changes over time. CD refers to a change in the class (concept) definitions of the data over time, i.e., a set of data instances has legitimate class labels at one time step, and the same set of data instances has different legitimate labels at another time step. Concepts may change according to various patterns (for example, suddenly or gradually), and if concepts are viewed as shapes in a representation space, then they can change their shape, size, and location \cite{b5}. CD concerns changes in the input as well as output sample spaces and distributions. There are many important aspects about CD to consider, such as how to detect it (sensitivity), how to determine the time point of CD (localization precision), how to determine the drift regions in the data, how to ensure confidence in CD detection (specificity), how to determine the type of CD, for example, true CD or false alarms, and what to do with the detected CD in relation to the learner knowledge \cite{b7}.

Although various studies attempt to address CD, the work is rooted in statistical foundations \cite{b1, b7, b10, b19.1} or mathematical algorithms \cite{b3, b8, b9, b14, b19.2}, not in systems theoretical foundations. In general, there is still a lack of formal theoretical frameworks or principled methods for resolving CD \cite{b2}.

\noindent \\ \textbf{Example III.1: Concept Drift in Healthcare Fraud}\\
\noindent Consider the healthcare fraud detection learning system defined in Example II.1. Fraudsters change their behavior, which is reflected in their transactions, claims, treatments, prescriptions, i.e., $X$, to remain undetected. This CD in the healthcare data and makes it harder to detect and identify fraud accurately with the predictive function $f$, and may even necessitate re-training or otherwise learning $f$ over time. Knowing this, fraudsters may execute various patterns such as alternating between different types of fraud $Y$, doing so gradually or suddenly, or they may commit fraud at random at infrequent points in time to make the CD difficult to address. In short, CD is complex and addressing it is not just about the solution method for learning $f$ but also heavily involves systems context.

\subsection{A Systems Theoretic Formulation}\label{AA}

An OL system is defined as a system that enables the sequential accumulation of knowledge within an inference system (learner) over time. The goal of the OL system is to update the knowledge base of the learner by optimizing the trade-off between learning as much value from the data as possible to obtain the most accurate learner whilst identifying when data instances become outdated, obsolete, and potentially misleading, which impedes the learner's learning potential to new data trends and potentially reduces model performance \cite{b6}. This phenomenon is called the stability-plasticity dilemma. In this dilemma, \textit{stability} entails retaining knowledge regarding the supposedly stationary underlying concepts, while \textit{plasticity} refers to forgetting some (or all) of the outdated acquired knowledge to facilitate the learning of the new concepts.

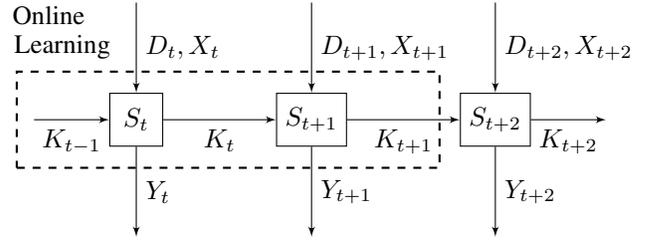
\begin{figure}
\centering
\begin{tikzpicture}[auto, node distance=2cm,>=latex']
\node [input, name=input] {};
\node [block, below=1.2cm of input] (system) {$S_t$};
\node [output, below=1.2cm of system] (output) {};

\node [block, right=1.5cm of system] (system1) {$S_{t+1}$};
\node [input, above=1.2cm of system1] (input1) {};
\node [output, below=1.2cm of system1] (output1) {};

\node [block, right=1.5cm of system1] (system2) {$S_{t+2}$};
\node [input, above=1.2cm of system2] (input2) {};
\node [output, below=1.2cm of system2] (output2) {};

\node [coordinate, left  =1cm of system] (input0) {};
\node [coordinate, right =1cm of system2] (output3) {};

\draw [draw,->] (input) -- node{$D_t,X_t$} (system);
\draw [->] (system) -- node{$Y_t$}(output);
\draw [->] (system) -- node[below]{$K_t$}(system1);

\draw [draw,->] (input1) -- node{$D_{t+1}, X_{t+1}$} (system1);
\draw [->] (system1) -- node{$Y_{t+1}$}(output1);
\draw [->] (system1) -- node[below]{$K_{t+1}$}(system2);

\draw [draw,->] (input2) -- node{$D_{t+2}, X_{t+2}$} (system2);
\draw [->] (system2) -- node{$Y_{t+2}$}(output2);

\draw [->] (input0) -- node[below]{$K_{t-1}$}(system);
\draw [->] (system2) -- node[below]{$K_{t+2}$}(output3);

\node[draw, thick, dashed, inner xsep=3.5em, inner ysep=0.8em, fit=(system) (system1)] (box){};

\node[above=0.1cm of box, inner sep=0pt](box0) at (box.north west) {};

\node[above right, inner sep=0pt](box1) at (box0.north west) {Learning};
\node[above=0.08cm of box1, inner sep=0pt](box2) at (box1.north west) {};
\node[above right, inner sep=0pt] at (box2.north west) {Online};

\end{tikzpicture}
\caption{OL from a systems theoretic perspective.}
\label{fig:OnlineL}
\end{figure}

Using these informal concepts, OL is formalized using ALST in Definition 1 below and illustrated in Figure \ref{fig:OnlineL}. Example III.2. illustrates OL systems in terms of the healthcare fraud detection system example.\\

\noindent \textbf{Definition 1: Online Learning Systems}\\
\noindent Given a learning system $S$ observed at two sequential points in time, i.e., $S_t = \{D_t, X_t, Y_t\}$ and $S_{t+1} = \{D_{t+1}, X_{t+1}, Y_{t+1}\}$ respectively, $S$ is an online learning system if it tries to update the knowledge $K_t$ of $S_t$ to improve the prediction performance of $S_{t+1}$, where knowledge at time $t$, $K_t$, is relation on data $D_t$ and parameters $\Theta_t$ at time $t$\footnote{This follows the conventional concept of knowledge in transfer learning, where knowledge is transferred in terms of samples of data, related feature representations, or learned model parameters \cite{cody1}.}.\\

\noindent \textbf{Example III.2: Online Learning System for Healthcare}\\
\noindent Consider the healthcare fraud detection learning system defined in Example II.1. To address CD, the learning system can become an OL system for healthcare fraud detection by using the knowledge $K_t$ learned by the learner $S_t$ at time step $t$ together with the previous observed claims data $X_t$ and its associated labels $Y_t$ to improve the learner to a new state $S_t$ with the goals of improving the prediction performance on newly observed claims data $X_{t+1}$ and better approximate the predictive function $f$.\\

The definition of OL systems can be further elaborated as follows: all OL systems are the union of systems with the stationarity assumption, called ``trivial OL systems'', and systems with the non-stationarity assumption, called ``non-trivial learning systems''. Stationarity in the input-output structure and behavior is not assumed in the definition for OL systems (Definition 1). But we can formalize a definition for non-trivial online learning systems as follows. \\

\noindent \textbf{Definition 2: Non-trivial Online Learning Systems} \\
\noindent Given a learning system $S$ observed at two sequential points in time, i.e. $S_t = \{D_t, X_t, Y_t\}$ and $S_{t+1} = \{D_{t+1}, X_{t+1}, Y_{t+1}\}$ respectively, non-trivial online learning systems try to update the knowledge $K_t$ of $S_t$ to improve the prediction performance of $S_{t+1}$, where $X_t \times Y_t \not = X_{t+1} \times Y_{t+1}$, or $P(X_t) \not = P(X_{t+1})$ or $P(Y_t|X_t) \not = P(Y_{t+1}|X_{t+1})$.\\

Non-trivial OL systems are closely related to transfer learning (TL) systems \cite{cody1, hpaper}. TL is concerned with the transfer of knowledge from source learning systems to a target learning system to help learning in the target. In the conventional ML literature, online transfer learning (OTL) is a concept concerned with  making predictions for target data which arrives in an online or sequential fashion, by training on offline labeled instances from a source data set \cite{OTL_Wu2, OTL_Wu1, b1, OL4_Zhoa1, OTL_Ge, OTL_Grubinger}. However, this ML concept for OTL is not comprehensive. For example, it does not concern the case where the source is also an OL system. 

Moreover, it underemphasizes the similarities between OL and TL. For example, OL can be modeled as successive TL where at each time step the source is $S_t$ and the target is $S_{t+1}$. Because (non-trivial) TL assumes a difference in $X$, $Y$, $P(X)$, or $P(Y|X)$ between sources and the target \cite{cody1}, such a successive TL system would also constitute a non-trivial OL system. Therefore, by focusing on non-trivial OL systems in this paper, any results ought to similarly apply to cases of repeated TL over time.

Traditionally, ML approaches focus on the decomposition of learning into the problem and the solution. In this paper, the use of system structure $\{X, Y \}$ and system behavior in terms of the joint distribution $P(X,Y)$ is proposed instead \cite{cody1}. This presented definition is more amenable to developing a systems theoretic framework for the design of OL systems. \\

\noindent \textbf{Example III.3: Structure and Behavior in Healthcare}\\
\noindent Consider the healthcare fraud detection OL system defined in Example III.2. The structure of the input sample spaces $X$ can vary over time $t$ when features are added or removed, i.e., $dim\{X_{t+1}\} \neq dim\{X_{t}\}$, in the claims data. The structure of the output sample spaces $Y$ can vary if the requirements for the learner predictions change (i.e., $Y$ changes from a binary fraud detection to a multi-class detection of fraud type). Even if the structure of the input and output spaces remains unchanged over time, the behavior in terms of the distribution of inputs $P(X)$ and outputs $P(Y|X)$ may vary over time.

\section{System Structure}
An important aspect of system structure for OL systems is the relationship between the learning system structures at successive time steps. These relationships can be captured by using mappings to structurally relate the learning systems over time. We focus on the structure of the input-output space of the predictive function $f$. 

For a learning system $S$ observed at two points in time, i.e., $S_t \subset \times \{D_t, X_t, Y_t\}$ and $S_{t+1} \subset \times \{D_{t+1}, X_{t+1}, Y_{t+1}\}$ respectively, the following are possible relationships between the input-output structures:

\begin{align}
    X_t = X_{t+1},  Y_t = Y_{t+1} \label{eqn:struc1} \\
    X_t \not = X_{t+1},  Y_t = Y_{t+1} \label{eqn:struc2} \\
    X_t = X_{t+1},  Y_t \not = Y_{t+1} \label{eqn:struc3}\\ 
    X_t \not = X_{t+1},  Y_t \not  = Y_{t+1} \label{eqn:struc4} 
\end{align}

When the sample spaces are equal or similar, as in Equation \ref{eqn:struc1}, it is said that $S_t$ and $S_{t+1}$ have \textit{homogeneous} system structures, i.e., identical system structures are observed, however, the same may not necessarily hold for the system behavior. Creating or assuming homogeneous system structures may be impractical, for even if a system is designed to have homogeneous structures, it can lose the shared structures due to degradation, maintenance, or rebuilds, which lead to changes in the input space. Another challenge is that proving the homogeneity of system structures may be theoretically infeasible. The concept of homomorphism can be used to address these challenges \cite{b21, b22, cody1}. Homomorphism is defined as follows.\\

\noindent \textbf{Definition 3: Homomorphism}\\
\noindent Two input-output systems $S^i = \{X^i, Y^i\}$ and $S^j = \{X^j, Y^j\}$ are homomorphic
if there exists a pair of maps,
\[\varrho : X^i \rightarrow X^j, \vartheta : Y^i\rightarrow Y^j\]
such that for all $x^i \in X^i, x^j\in X^j$ and $ y^i \in Y^i, y^j \in Y^j$, $\varrho(x^i) = x^j$ and $\vartheta(y^i) = y^j$.\\

\noindent Alternatively, $S^j$ is called a \textit{homomorphic image} of $S^i$ if there exists a mapping from the system structure of $S^i$ onto that of $S^j$. 

Equations \ref{eqn:struc2} and \ref{eqn:struc3} illustrate partially heterogeneous system structures and \ref{eqn:struc4} describes fully heterogeneous system structures. In these cases homogeneous structures can be recovered by using homomorphic mappings. One can ensure that a system $S^i$ has a homogeneous structure with another system $S^j$ by finding the appropriate $\varrho$ and $\vartheta $ such that the sample spaces of $S^i$ can be mapped onto the sample spaces of $S^j$. An alternative approach could be to map each sample space to a different input-output space that is neither identical to $S^i$ nor $S^j$ input-output spaces.\\

\noindent \textbf{Example IV.1: Healthcare Fraud OL System}\\
\noindent Consider a healthcare fraud detection learning system $S$ observed at two sequential points in time, i.e., $S_t = \{D_t, X_t, Y_t\}$ and $S_{t+1} = \{D_{t+1}, X_{t+1}, Y_{t+1}\}$, respectively, where $X_t = X_{t+1}$ and $Y_t \not = Y_{t+1}$. Let $Y_t = \{0,1\}$ be a binary indicator of fraud where ``0" represents normal and ``1" represents fraudulent behavior. Let $Y_{t+1} = \{0,1,2,3,4\}$ be a multi-class indicator where ``0" represents normal behavior and values ``1",``2",``3" and ``4" represent specific types of fraud, for example over-billing, upcoding, duplicate transactions and unnecessary auxiliary services frauds. 
\noindent A homomorphic mapping can be defined by defining $\varrho$ as the identity mapping, i.e., $\varrho(y_t) = y_{t+1}$, and $\vartheta$ such that

\begin{equation}
  \vartheta(y_{t+1}) =
    \begin{cases}
      0 & \text{if $y_{t+1} = 0$}\\
      1 & \text{if $y_{t+1} = 1,2,3,4$}\\
    \end{cases}       
\end{equation}

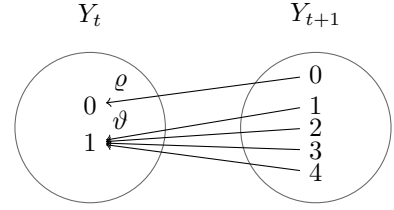
\begin{figure}[t]
\centering
\begin{tikzpicture}
    \filldraw[fill=white!20, draw=black!60] (-1.5,0) circle (1cm);
    \filldraw[fill=white!20, draw=black!60] (1.5,0) circle (1cm);

    \node at (-1.5,1.5) {$Y_t$};
    \node at (1.5,1.5) {$Y_{t+1}$};

    \node (x1) at (-1.5,0.3) {$0$};
    \node (x2) at (-1.5,-0.2) {$1$};

    \node (y1) at (1.5,0.7) {$0$};
    \node (y2) at (1.5,0.3) {$1$};
    \node (y3) at (1.5,-0.0) {$2$};
    \node (y4) at (1.5,-0.3) {$3$};
    \node (y5) at (1.5,-0.6) {$4$};

    \draw[->] (y1) -- (x1);
    \draw[->] (y2) -- (x2);
    \draw[->] (y3) -- (x2);
    \draw[->] (y4) -- (x2);
    \draw[->] (y5) -- (x2);

    \node (rho) at (-1.1,0.6) {$\varrho$};
    \node (rho) at (-1.1,0.1) {$\vartheta$};

\end{tikzpicture}
\caption{A homomorphic map from $Y_t = \{0,1\}$ to $Y_{t+1} = \{0,1,2,3,4\}$}
\label{fig:map}
\end{figure}

\noindent These mapping functions enables the mapping of $S_t$ onto $S_{t+1}$ and therefor the OL system between the two system observations can assume a homogeneous system structure. Figure \ref{fig:map} illustrates the mapping functions. \\

To determine the similarity of system structures by using the entire sample spaces $\{X_t, Y_t\}$ and $\{X_{t+1}, Y_{t+1}\}$ is an impractical endeavor since it offers too broad a scope. However, subspaces could be identified and used instead. Suppose $X_t \not = X_{t+1}$, but that they have shared subspaces, i.e., subspaces that are equal or homogeneous. Then concepts of homogeneous, partially heterogeneous, and fully heterogeneous system structures can be extended to homogeneous, partially heterogeneous, and fully heterogeneous system substructures using parallel definitions \cite{cody3}. Shared substructures could be used to infer relationships between the system structures. When there are no homogeneous substructures but partially or fully heterogeneous substructures instead, there may be some shared latent or unobserved features in the data that could be used to infer relationships between the system structures. The challenge becomes how to discover this latent data and how to use it to relate system structures. A potential solution could be the utilization of subject matter experts which exist externally to the learning system and may be both difficult and expensive to acquire \cite{cody1}.\\

\noindent \textbf{Example IV.2: Substructures in Healthcare Fraud}\\
\noindent Consider a healthcare fraud detection learning system $S$ observed at two sequential points in time, i.e., $S_t = \{D_t, X_t, Y_t\}$ and $S_{t+1} = \{D_{t+1}, X_{t+1}, Y_{t+1}\}$, respectively. Suppose the claims data $X$ observed at time $t$ contains features such as provider identifier, address details, and specialty; and billing details such as date, patient details, and claim amounts. Suppose the claims data observed at time $t+1$ has fewer features, i.e., the provider address details such as country, state or province, city and exact street address have been omitted. Hence, $X_t \not = X_{t+1}$. Also assume that $Y_t = Y_{t+1}$. 
The homogeneous substructures consist of all the features except for the provider address details, and they can be used to maintain the system knowledge. Additionally, predictions $Y_{t+1}$ can be made about the observed fraudulent behavior by using the features of $X_t$ and inferring a mapping from $X_{t+1}$ onto $X_t$ using the homogeneous substructures. Now, suppose that $X_{t+1}$ contains procedures, supplies, products, and service codes. Now there are no homogeneous substructures present and inferring a mapping between $X_t$ and $X_{t+1}$ is unlikely without the help of external knowledge. Latent details might be present in some of the observed data, however, they are not likely to be found through the use of the system alone. Knowledge external to $S_t$ and $S_{t+1}$, such as expert knowledge from a health insurance firm, could be utilized to relate the two system structures.


\section{System Behavior}
As mentioned before, the primary system behavior of interest is defined by the probability measures on the predictive function $f : X \rightarrow Y$ or alternatively, the marginal and conditional distributions of the joint distribution $P(X,Y)$. In this section two aspects of OL system behavior will be discussed, namely CD and the knowledge base. 

\subsection{Concept Drift}
CD is the occurrence where the underlying data distribution changes over time. Characteristics of CD include the pattern of drift, or rate of change, and the type of drift. The types of drift are related to the system behavior as follows. 

When there is a change in the underlying data distribution of the input, it is called virtual drift, and it is typically discovered through unsupervised learning. Virtual drift refers to the non-stationarity of the marginal distribution $P(X)$ or the joint distribution $P(X,Y)$ (mathematically, the two distributions are the same). The changes in the distribution may be due to an incomplete or partial feature representation of the current data distribution, hence there could be changes in the number of attributes. While the predictive function that characterizes the relation between $x_t$ and $y_t$ may stay the same for virtual drift, the change in distribution places different importance on different parts of the feature space for the algorithm to digest \cite{b10, b15}.  Virtual drift is also called covariate drift, sampling shift, temporary drift, or feature change. Virtual drift specifies a difference between input behavior, and the output behavior may or may not remain equal, however, it is common that when the marginal distribution changes, the posterior distribution changes (real drift occurs) since the input set conditioning on the posterior distribution has changed.  

When there is a change in the target concept, it is called real drift, and it is typically identified through supervised learning. Real drift refers to the non-stationarity of the posterior distribution $P(Y|X)$, which may be caused by a change in the class boundary (the number of classes) or the class conditional probabilities (likelihood) $P(X|Y)$ \cite{b15, b23}. Real drift could occur with or without any change in the marginal distribution $P(X)$. Real drift is also called class drift, concept shift, or conditional change.

Note that the behavioral difference in the posterior distributions can be induced by a structural difference in the inputs. Usually, when the underlying distributions change, the change will be evident in the observed sample spaces.

\noindent \\ \textbf{Example V.1: Concept Drift in Healthcare Fraud}\\
\noindent Consider the healthcare fraud detection OL system defined in Example III.2. In the case of virtual drift, fraudsters would change their behavior $P(X)$, for example instead of overbilling for a specific procedure or treatment they charge for multiple unnecessary auxiliary services which still leads to the claim being more expensive than it should be hence they are still classified as fraudulent. Even though the marginal distribution $P(X)$ changes, the OL system still focuses on expensive claims to indicate fraud, i.e., $P(Y|X)$ may not change significantly. In the case of real drift, fraudsters would for example move away from overbilling or excessive charges on claims and move towards inappropriate or over-providing medication. In this case both the marginal distribution $P(X)$ and posterior distributions $P(Y|X)$ change since the OL system shifts its focus to outliers in the relationship between diagnoses and their appropriate treatments.

\subsection{Knowledge Base}
As mentioned before, OL shares similarities with TL and some TL concepts can translate well to OL. Inspired by TL, there are three main approaches to update the knowledge of a learning system between two time steps, $t$ and $t+1$ respectively, namely:
\begin{enumerate}
    \item Using input-output pairs observed in time step $t$
    \item Using the parameters of the learner $S_t$, which has been learned from the data $D_t$
    \item Determining the features that minimize the divergence between the two sequential domains \cite{cody3}.
\end{enumerate}

The last approach relates to transferring system structure since it requires finding and sharing appropriate feature representations of the domains, whereas the first two relate to system behavior. The input-output pairs provide empirical evidence of system behavior since they consist of joint and marginal instances of the sample spaces $X$ and $Y$. The learner offers an approximation or model of the system behavior since it was trained on the observed data to learn the parameters of the predictive function, which maps $X$ to $Y$. 

\section{Discussion}

Structure and behavior provide another perspective to OL systems which enables broader investigation and expansion. Structural considerations focus on the structural relationship between $S_t$ and $S_{t+1}$ and the value of both homogeneous and heterogeneous structures for OL systems. Behavioral considerations focus on both the closeness and difference of the probability distributions between $S_t$ and $S_{t+1}$ and the complexity of the knowledge base used to update the predictive function for the OL system as time progresses. These perspectives provide first principles for the top-down design and analysis of OL systems.

The primary benefit of a systems theoretic perspective on OL is the analysis of the relationships between system structure and system behavior, and between the system behavior and the observed data distributions. Important questions arise regarding system structure and behavior when designing OL systems, for example (1) what is the relationship between the system structure and system behavior for a specific system of interest, (2) what structural similarities and dissimilarities exist, (3) how complex is the behavior of interest to learn, and (4) how variable is the behavior over time. Complexity and variability are viewed as subjective characteristics relative to system structure and system behavior \cite{cody3}. 

CD is a critical challenge faced by OL, as it leads to much complexity and variability in OL systems, however it is difficult to identify and therefor control. It manifests in various forms, for example the observed pattern(s) of drift (rate of change) and the type of drift. CD patterns include gradual, incremental or sudden drifts; concepts may reappear at various rates; or there could be noise or sporadic outliers in the data \cite{b8}. Each pattern of CD has a unique complexity, variability, and relationship with the system structure. The complexity of the behavior of a system with a structure $\{X, Y\}$ may be different than with a structure $\{X^`, Y\}$, potentially due to virtual drift. The same may be said regarding the variability between systems. The system behavior of a system observed over time may seem different under a particular homogeneous structure but nearly identical under another. For a fixed input-output space, control over the joint distribution $P(X, Y)$ via the design of the system and its related procedures can also play a role.\\

\noindent \textbf{Example VI.1}\\
Consider the complexity, variability, and scale of a healthcare fraud detection system such as that described in Example III.2. Various factors and variables are at play. Fraud can be committed by patients, insurers, and/or providers. Numerous information systems capture, process, manage, integrate, and store healthcare data at various locations, i.e., providers and insurers. Thousands of healthcare codes for procedures, supplies, products, and services are captured and processed. Billions of healthcare claims are created annually. Lastly, and perhaps most importantly, labels for healthcare fraud detection are extremely scarce. The quantity and quality of sample spaces provided to the learning system acutely affect the complexity and variability of the OL system. If the OL system uses limited sample spaces to learn from, then the system behavior may become complex to ensure that the relationships are appropriately captured. The way in which the inputs are selected and transformed, i.e., the choice of system structure from the set of homogeneous system structures, has an impact as well. It is difficult to select appropriate healthcare features and to transform them appropriately to enhance fraud detection, given the wealth of healthcare features that are captured. System behavior may appear less variable after a simple transformation, such as standardization of the inputs or the application of semantic embedding algorithms prior to learning. The output of the OL system can have an impact as well. A multi-class indicator may be more difficult to learn than a binary indicator. Clearly these factors shape the viable and optimal solution space for OL designs, and many are assessed at the systems level, either in terms of the OL system itself, or in terms of its relationship with other systems, users, and environments.\\
\\

The design decisions regarding the system structure, i.e. the observed input and output spaces $X, Y$ over time may occur long before algorithm design, yet the success of algorithm design is heavily determined by these choices. The complexity and the variability of the system behavior, i.e. $P(X, Y)$ and by implication CD over time, also plays a crucial role in OL success. The control over the overall system complexity and variability is vital to ensure trustworthy OL systems, however it may be unavailable. 

Unlike the current OL literature, algorithm design and implementation is not the primary focus the presented systems theoretic formulation of OL herein. Instead this work offers a novel decomposition of OL systems in terms of system structure and behavior, which enables the identification of key design parameters which can be utilized to understand and control complexity and variability within these systems, which has the potential to enhance OL system robustness and reliability. 

\section{Conclusion}
The work in this paper advocates for the development of systems theoretic frameworks to guide these ML applications. OL is an active field of research and has been widely explored, however, in general, there still lacks formal theoretical frameworks or principled methods for modeling OL systems and resolving CD. The presented work provides a foundation for formulating OL in the context of systems theory. The presented formalism aids in approaching OL as a top-down design problem by offering an evaluation of both the system structure and system behavior. 

Important aspects of OL systems were discussed, namely the homogeneity and heterogeneity of system structures, the relationship of CD to system behavior, and the process of relating the observed input-output spaces so that knowledge of the relationship between inputs and outputs can be maintained within the learning system as time progresses.

CD is a critical challenge in OL. Future work will aim to delve deeper into key aspects of CD and to formalize definitions of characteristics such as drift localization, drift sensitivity, drift specificity, and actual versus counterfeit drift. The framework can be further expanded by formalizing broader notions such as stability learning and plasticity learning in accordance with the stability-plasticity dilemma. If these aspects can be formalized and integrated, then the framework will constitute a formal systems theory of OL.


\bibliographystyle{ieeetr}
\bibliography{biblio}

\end{document}